\pdfoutput=1
\documentclass[11pt]{article}

\usepackage[]{acl}

\usepackage{times}
\usepackage{latexsym}

\usepackage[T1]{fontenc}
\usepackage[utf8]{inputenc}

\usepackage{microtype}
\usepackage{amssymb}
\usepackage{amsopn}
\usepackage{subcaption}
\usepackage{graphicx}
\usepackage{multirow}
\usepackage{makecell}
\usepackage[english]{babel}
\usepackage{bm}
\usepackage{array}
\usepackage{setspace}
\usepackage{CJKutf8}
\usepackage{algorithmicx}  
\usepackage{algpseudocode}  
\usepackage{amsmath}  
\usepackage{color,xcolor}
\usepackage{algorithm} %format of the algorithm 

\usepackage{pifont}
\usepackage{bm}
\usepackage{amsfonts}

\usepackage{graphicx}
\usepackage[normalem]{ulem}
\usepackage{multirow}
\usepackage{amsmath}
\usepackage{url}

\usepackage{tikz}
\usepackage{xcolor}
\usepackage{tcolorbox}
\usepackage{color,xcolor}

\usepackage{amssymb}
\usepackage{amsopn}
\usepackage{graphicx}
\usepackage{multirow}
\usepackage[english]{babel}
\usepackage{bm}
\usepackage{array}
\usepackage{setspace}
\usepackage{todonotes}

\usepackage{xspace}
\usepackage{amsfonts}
\usepackage{array,multirow}
\usepackage{pgfplots}
\usepackage{tikz}
\usepackage{verbatim}
\usepackage{arydshln}
\usepackage{booktabs}
\definecolor{ugreen}{rgb}{0,0.5,0}
\definecolor{mygreen}{RGB}{58,127,88}
\definecolor{iyellow}{RGB}{255,250,205}
\definecolor{ipurple}{RGB}{230,230,250}
\definecolor{myred}{RGB}{160,52,52} %238，44，44
\definecolor{myblue}{RGB}{30,144,255}
\definecolor{myorange}{RGB}{255,127,80}
\definecolor{mypurple}{RGB}{255,20,147}

\title{Contrastive Learning with Prompt-derived Virtual Semantic Prototypes for Unsupervised Sentence Embedding}

\author{Jiali Zeng$^{1}$, \  Yongjing Yin$^{2}\thanks{\ \ Corresponding author.}$, \ Yufan Jiang$^{1}$, \ Shuangzhi Wu$^{1}$, \ Yunbo Cao$^{1}$ \\
$^{1}$Tencent Cloud Xiaowei, Beijing, China \\
$^{2}$Westlake University, Zhejiang, China\\
{\tt \{lemonzeng,garyyfjiang,frostwu,yunbocao\}@tencent.com} \\
{\tt yinyongjing@westlake.edu.cn} \\ 
}

\begin{document}
% \begin{CJK*}{UTF8}{gbsn}
\maketitle
\begin{abstract}

% Unsupervised sentence embedding aims to obtain the most appropriate embedding for a sentence to reflect its semantic.
Contrastive learning has become a new paradigm for unsupervised sentence embeddings.
Previous studies focus on instance-wise contrastive learning, attempting to construct positive pairs with textual data augmentation.
In this paper, we propose a novel {\bf Con}trastive learning method with {\bf P}rompt-derived {\bf V}irtual semantic {\bf P}rototypes (ConPVP). 
Specifically, with the help of prompts, we construct virtual semantic prototypes to each instance, and derive negative prototypes by using the negative form of the prompts.
Using a prototypical contrastive loss, we enforce the anchor sentence embedding to be close to its corresponding semantic prototypes, and far apart from the negative prototypes as well as the prototypes of other sentences.
Extensive experimental results on semantic textual similarity, transfer, and clustering tasks demonstrate the effectiveness of our proposed model compared to strong baselines.
Code is available at https://github.com/lemon0830/promptCSE.
% By comparing with similar and opposite semantic examples, XXX.
% the model can effectively perceive the semantic changes cased by small perturbations.
% Empirical results show that our approach yields substantial improvements on a range of semantic textual similarity tasks compared with strong baselines.
% In addition, we conduct extensive analysis and apply the learned sentence embedding to transfer and clustering tasks to confirm the effectiveness and robustness of ConPVP.

\end{abstract}

\section{Introduction}

High-quality sentence embeddings can boost the performance of pre-trained language models (PLMs) on many downstream tasks \cite{Kiros-etal-2015-skip-thought-vectors, Logeswaran-etal-2018-an-efficient, Reimers-etal-2019-sentencebert}.
Recent research focuses on learning sentence embeddings in an unsupervised manner due to lack of large scale labeled data \cite{Hill-etal-2016-learning, Pagliardini-etal-2018-unsupervised, Wang-etal-2021-TSDAE}.
% In particular, contrastive learning \cite{} and has proven remarkably successful.
% Unsupervised sentence embedding aims to obtain the most appropriate embedding for a sentence to reflect its semantic.
Among these methods, contrastive learning has been extensively explored and achieved remarkable success \cite{gao-etal-2021-simcse,wu-etal-2021-esimcse,yan-etal-2021-consert, giorgi-etal-2021-declutr}.
% Specifically, 
% itnotions of similar (positive) and dissimilar (negative) pairs of samples, and uses an InfoNCE loss \cite{Oord-etal-2018-representation, He-etal-2020-momentumcl} to map representations of positive pairs gathered and those of negative pairs farther apart，
% (e.g., swapping, inserting, and deleting words) 
Specifically, most of them construct a positive pair by operating various textual data augmentation methods, while regard two independent sentences sampled uniformly from the training data as a negative pair.
% focus on generating positive samples using diverse data augmentation, and use other independent sentences sampled uniformly from the training data as negative samples, 
In spite of effectiveness in easing the anisotropy problem, such instance-wise optimization leads to a locally smooth embedding space, and ignores semantic relevance to some extent \cite{li-etal-2020-prototypical}. 
% making each instance locally smooth ignoring semantic relevance.
% the contrastive learning based methods 
% first generate positive samples by textual data augmentation and regard the sampled sentences as negative samples, then 
% use an instance-wise contrastive loss \cite{Oord-etal-2018-representation, He-etal-2020-momentumcl} to gather positive samples based on data augmentation and push apart negative pairs that sampled from the training data, however, leading to a local smoothness problem \cite{li-etal-2020-prototypical}. 
% However, such contrastive mechanism ignoring intrinsic semantic structure tends to make each instance have a similar representation to its augmentation (i.e., local smoothness issue) \cite{li-etal-2020-prototypical}. 
Moreover, due to the discrete nature of language, data augmentation can change sentence semantics significantly, and thus a positive sample are possibly turned into a negative one \cite{Wang-etal-2021-CLINE}.

\begin{figure}[!t]
\centering
\includegraphics[width=1.0\linewidth]{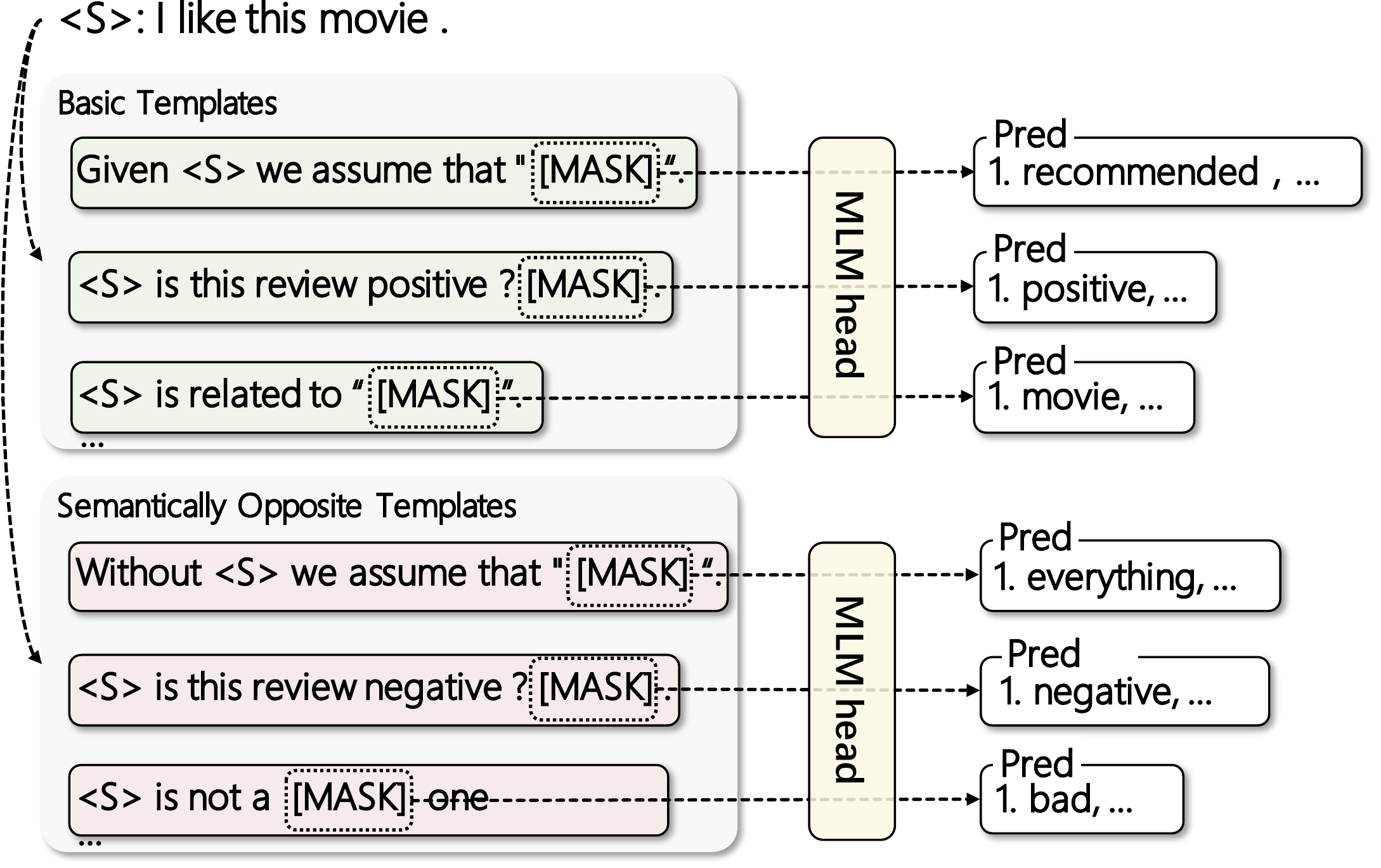}
\caption{
Paradigm of prompt learning.
}
\label{fig_intro1}
\end{figure}

To alleviate the issues, we introduce the idea of prototypical contrastive learning to unsupervised sentence embeddings learning, which is proven effective to learn structural visual embedding space \cite{li-etal-2020-prototypical,Caron-etal-2020-contrasting-cluster-assignments}.
The motivation lies in that when encoding sentences into the embedding space, 
% the semantic structures should be implicitly modeled, i.e., 
the sentences with similar semantics cluster together around the corresponding prototype.
% Prototype is defined as ``a representative embedding for a group of semantically similar instances'' \cite{}.
% Recent studies of self-supervised learning in  have shown that the prototypical contrastive learning successfully learns a structural embedding space \cite{}.
% Recent studies of prototypical contrastive learning in computer vision 
% In CVs, the prototypical contrastive learning assigns several prototypes discovered by clustering to each image, and enforce the image embedding to be closer to the assigned prototypes.
% In this paper, we introduce the idea of prototypical contrastive learning to unsupervised sentence embeddings learning.
Nevertheless, the acquisition of prototypes is inefficient if we directly apply the clustering algorithms used in \cite{li-etal-2020-prototypical,Caron-etal-2020-contrasting-cluster-assignments},
% Nevertheless, it is inefficient to if directly apply the clustering algorithms 
% used in \cite{li-etal-2020-prototypical,Caron-etal-2020-contrasting-cluster-assignments} to acquire prototypes, 
due to the requirement of extra forward pass over the training set or weak correlation with semantics \cite{Caron-etal-2020-contrasting-cluster-assignments}.
% careful tuning of hyper-parameters.
This makes us wonder {\bf whether there exists a dedicated method of mining the semantic prototypes for sentence embeddings especially based on PLMs?}
To answer this question, we attempt to think from the perspective of prompt learning \cite{Brown-etal-2020-languagemodelasfewshotlearner}.
% for sentence-level NLP tasks. 
Intuitively, on sentence-level NLP tasks such as classification, the neural models encode and map each input to a corresponding semantic prototype in the embedding space.
% The encoder encodes the task-specific semantic information and the encoded embedding serves to conduct discrimination.
% We argue that 
% In particular, the discrimination is to cluster data into different semantic prototypes.
For example, the sentiment analysis models divide instances into semantic prototypes related to sentiment polarity.
% Different tasks discrimination are to assign data into different semantic clusters of different perspective, where the semantic clusters can be viewed as prototypes.
% : ``Positive'' or ``Negative''.
% , which can be viewed as semantic prototypes.
% Therefore, 
% we can directly treat the categories of different tasks as semantic prototypes of different perspective.
% However, 
% the space of textual label words for each task is somehow infinite due to the diversity of natural language.
% For example, the label words for semantic classification can be ``Happy'' or ``Sad'' instead.
% Besides, 
% without task annotated corpus or pretrained task-specific models, we can not obtain corresponding label for each training instance. 
% Specifically, 
% We treat the sentence embedding task as a (masked) language modeling problem, where an input sentence is concatenated with a {\it Continuous Prompt} and a {\it Representation Token} (e.g. [MASK] token) and fed to PLMs to obtain the hidden state of the {\it Representation Token} as sentence embedding.
% Specifically, 
In addition, 
% due to the great power of the general-purpose PLMs, 
the sentence-level tasks can be solved by providing task-specific prompts to PLMs as a condition even without any fine-tuning \cite{Brown-etal-2020-languagemodelasfewshotlearner,Sanh-etal-2021-multiprompt, Wei-etal-2021-fintuned-language-models-are-zero-shot-learners}.
% PLMs are demonstrated good at performing sentence-level tasks like classification using well-designed prompts, with merely few parameter update \cite{} or even without any fine-tuning \cite{Brown-etal-2020-languagemodelasfewshotlearner,add more}.
% For prompt-based PLMs, 
% the knowledge acquired by PLMs can be better utilized by the mask mechanism.
% ``re-using'' the mask mechanism.
As illustrated in Figure \ref{fig_intro1}, PLMs can directly generate reasonable label words (e.g., ``positive'') for a sentence <$S$> by answering the query of the ``[MASK]'' token, when fed a prompt-wrapped sequence (e.g., ``<S> is this review positive ? [MASK]'').
% (e.g., ``I like this movie''), a task-specific template (e.g., ``<S> is this review positive ? ''), and .
% To specify, 
% by wrapping input into a template and using a verbalizer which constructs a mapping between label space and label word space, prompt tuning can achieve excellent results in zero-shot and few-shot scenarios.
% \cite{} stated prompting impacts the pretrained models' efficiency and is often worth hundreds of data points on average across classification tasks.
% a task-specific template acts as a description in natural language of the task.
% To specify,
% equip a sentence with a task-specific template and a placeholder (i.e., ``[MASK]'' token), and then feed it into the PLMs can derive a reasonable label word (textual response).
% As illustrate in Figure \ref{fig_intro}(a), 
% given sentence ``I like this movie'', 
% we concatenate it with different task-specific templates and feed them into {\it BERT-base},
% ``Positive'', ``Recommended'' and ``Movie'' can be derived with 
% the final hidden states of the ``[MASK]'' token.
Thus, we argue that the representations of the ``[MASK]'' token derived by task-specific templates can be viewed as virtual semantic prototypes, which can be obtained without using label information \cite{Lan-etal-2021-CLLD} and clustering algorithms \cite{li-etal-2020-prototypical, Caron-etal-2020-contrasting-cluster-assignments}.
Besides the commonly-used templates, we
% construct another set of templates by 
manually convert each template to its negation, and use them to induce negative prototypes.
% having the opposite semantics.
% semantically opposite to its prototypical embedding. 
% simply convert the semantics of the basic prompt.
Back to Figure \ref{fig_intro1}, 
% by changing the word ``given'' to the word ``without'',
with the template ``<S> is not a [MASK] one'', the word ``bad'' can be derived from PLMs.

% Therefore, the representations of the ``[MASK]'' token derived by semantically opposite templates can be viewed as negative prototypes for each input sample.
% Thus, the conv

\begin{figure}[!t]
\centering
\includegraphics[width=1.0\linewidth]{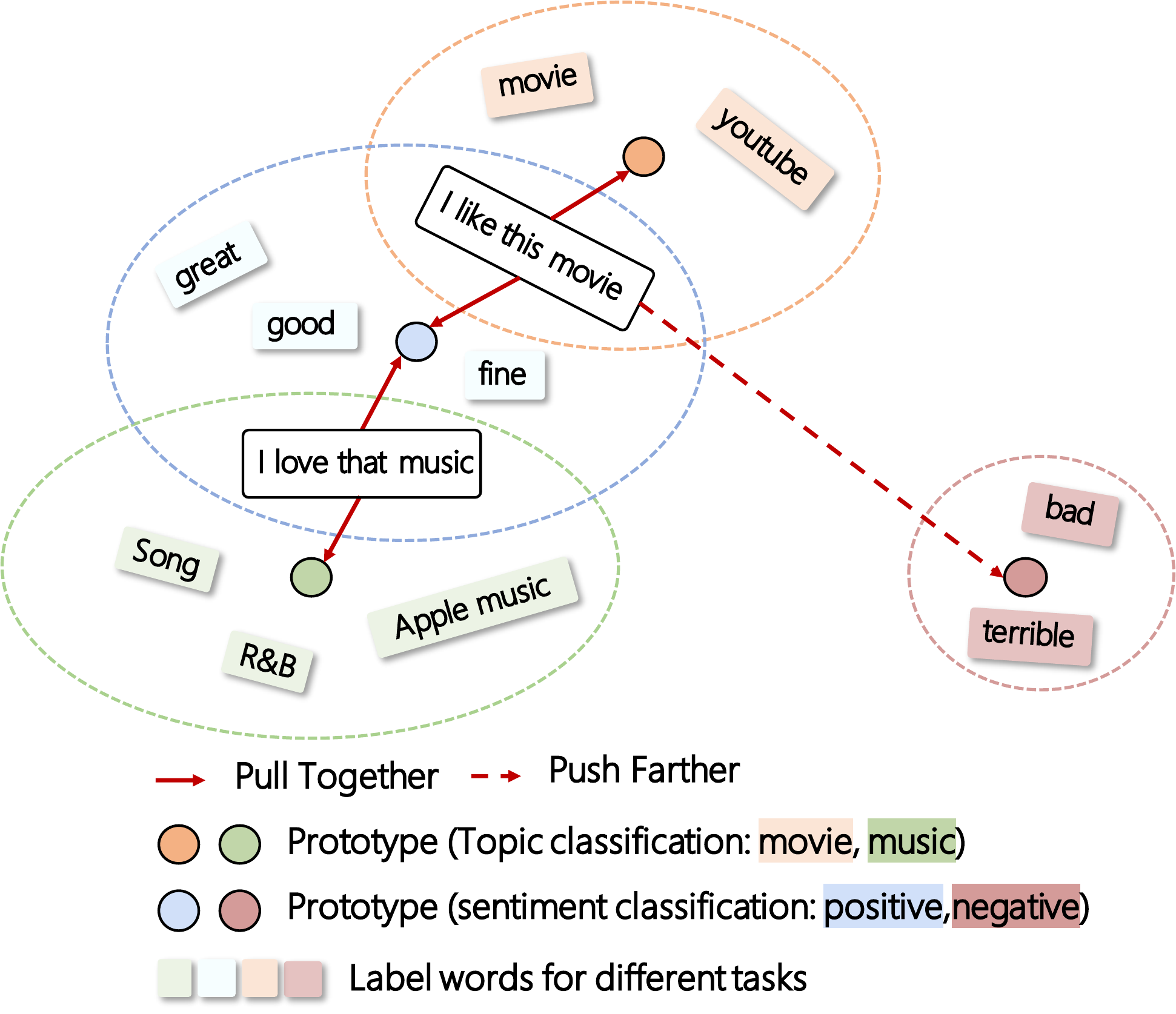}
\caption{
Illustration of contrastive learning with prompt-derived virtual semantic prototypes.
% Each instance is assigned to multiple prototypes induced by different prompt-based NLP tasks.
% ConPVP learns an embedding space which encodes the semantic structure of sentence.
}
\label{fig_intro2}
\end{figure}

% Driven by the above observations,  
% claims, in this paper, 
% we propose a prototypical {\bf Con}trastive learning with {\bf P}rompt-derived {\bf V}irtual Semantic {\bf P}rototypes for unsupervised sentence embedding, coined ConPVP, implicitly encoding semantic structure into the embedding space.
% derive {\bf Virtual Semantic Prototypes} by conducting multiple downstream task-specific prompt predictions without the need of additional labeled task-specific corpora and pre-training.
% First, 
% In particular, 
% we reformulate the sentence embedding task as the mask language model task to avoid embedding biases \cite{jiang2022promptbert}.
% Specifically, we feed an input sentence to PLMs after being concatenated with a discrete (or continuous) prompt and a ``[MASK]'' token, and obtain the hidden state of the ``[MASK]'' token as the sentence embedding.
% Then, 
% In a nutshell
% Based on the above insights, 
In this paper, we propose ConPVP ({\bf Con}trastive learning with {\bf P}rompt-derived {\bf V}irtual Semantic {\bf P}rototypes) (ConPVP) for unsupervised sentence representation learning.
Specifically, given an input sentence, we generate the positive and negative prototypical embeddings by using a task-specific template and its negative counterpart, respectively.
% and then 
% each sentence is assigned to multiple semantic prototypes derived by conducting task prompt predictions.
% We integrate the sentence embeddings and virtual prototype embeddings with a contrastive loss, 
We use the contrastive loss to enforce the sentence embedding to be close to its positive prototype, and far apart from the negative prototype as well as the prototypes of other sentences.
As illustrated in Figure \ref{fig_intro2}, 
% encoding the semantic structure into the embedding space, 
the issue of local smoothness
% of sentence representations 
can be alleviated by exploiting the semantic regularization induced by task-specific prompts, and the sentences with similar semantics are closer.
% Compared to the recent contrastive sentence embeddings methods, our proposed ConPVP has the following advantage: 
% \begin{itemize}
%     \item Unlike conventional contrastive loss that only uses instance-level supervision, our objective exploit the semantic regularization induced by task-specific prompts, encoding the semantic structure into the embedding space.
%     % ConPVP encodes the semantic structure discovered by task-specific prompts into the embedding space;
%     \item We push the anchor embedding far from the negative prototype embeddings to avoid undesirably push the instance representations sharing similar semantics apart. 
% \end{itemize}
% Unlike the conventional contrastive methods that only uses instance-level supervision, 
% our ConPVP exploits the semantic regularization induced by task-specific prompts, encoding the semantic structure into the embedding space.
% ConPVP encodes the semantic structure discovered by task-specific prompts into the embedding space;
% We push the anchor embedding far from the negative prototype embeddings to avoid undesirably push the instance representations sharing similar semantics apart. 
% Since any two superficially different sentences are regarded as a negative pair, 
% the representations of the instances sharing similar semantics are undesirably pushed apart \cite{wang2022sncse}.
We empirically evaluate our proposed ConPVP on a range of semantic textual similarity tasks, and the experimental results show the substantial improvements compared with strong baselines.
Further,the extensive analysis and applications to transfer and clustering tasks confirm the effectiveness and robustness of our ConPVP.

\section{Related Work}

\subsection{Prototypical Contrastive Learning}
Recently, 
% cluster-based contrastive method, represented by 
prototypical contrastive learning has shown its power in computer vision \cite{li-etal-2020-prototypical, Caron-etal-2020-contrasting-cluster-assignments, Sharma-etal-2020-clustering} and NLP tasks \cite{Wei-etal-2022-eliciting, Ding-etal-2021-prototypical}, which discover the underlying semantic structure by clustering the learned embeddings.
% In practice, we can find prototypes by performing clustering on the embeddings.
% encode the semantic structure into the representation space in computer vision.
% are proposed to capture high-level semantics information in computer vision.
% Taking inspiration from the idea, 
Compared with them, we propose a more efficient and dedicated method to find prototypes for sentence embeddings, without using clustering algorithms or label information.
% which is helpful for encoding semantic structures into the embedding space.
% finding truly semantic similar samples. 
To the best of our knowledge, we are the first to explore the prototypical contrastive learning in unsupervised sentence representation learning.

\subsection{Prompt-based Learning}
Prompt-based Learning has become a new paradigm in NLP, bridging the gap between pretraining tasks and downstream tasks \cite{Brown-etal-2020-languagemodelasfewshotlearner,schick-schutze-2021-exploiting, Sanh-etal-2021-multiprompt}. 
It reformulates various NLP tasks as cloze-style questions, and by doing so,
% and makes predictions using 
the knowledge stored in PLMs can be fully exploited, making PLMs achieve impressive performance in few-shot and zero-shot settings.
Along this research line, various types of prompts are explored including discrete and continuous prompts \cite{gao-etal-2021-making,shin-etal-2020-autoprompt,hu2021knowledgeable,liu2021gpt, cui-etal-2021-template,si2021generating,li-liang-2021-prefix,schick-schutze-2021-shot}.
In this work, we exploit prompts of different downstream tasks to assign various virtual semantic prototypes to each instance.
% Unlike these studies,  
% Our method 
% In addition, tasks that utilize prompts are extended from text classification task to more NLP tasks, like information extraction \cite{cui-etal-2021-template,si2021generating} and text generation task\cite{li-liang-2021-prefix,schick-schutze-2021-shot}.

\subsection{Unsupervised Sentence Embedding}
% Directly encoding sentence embedding from pretrained language model results in poor performance due to anisotropy phenomenon, 
Unsupervised learning has been used to improve the sentence embedding learning \cite{reimers-gurevych-2019-sentence,li-etal-2020-sentence,su2021whitening,zhang-etal-2020-unsupervised}, and contrastive learning has attracted extensive attention due to the promising performance \cite{gao-etal-2021-simcse, giorgi-etal-2021-declutr,wu2020clear,yan-etal-2021-consert,meng2021coco,carlsson2020semantic}.
% Most of them construct a positive pair by operating various textual data augmentation methods, while regard two independent sentences as a negative pair. 
% \cite{yan-etal-2021-consert,gao-etal-2021-simcse}.
% However, it has been pointed out that such instance-wise contrastive method with simple augmentation strategy will lead to sentence representation suppression.
% For example, 
\citet{wu-etal-2021-esimcse} augment positive pairs with word repetition and introduce a momentum encoder for negative pairs.
% to contrastive learning.
% \citet{jiang2022promptbert} reformulate the sentence embedding task as the mask language task to avoid the embedding biases.
\citet{wang2022sncse} use soft negative samples which have highly similar textual but opposite meaning to the input sentence.
\citet{jiang2022promptbert} use a discrete template to obtain sentence embeddings.
Unlike these studies, we introduce prototypical contrastive learning and implicitly encode semantic structure induced by task-specific prompts into the embedding space, enhancing PLMs' ability of modeling semantic similarity. 
Furthermore, our prototypical contrastive loss is orthogonal to the instance-wise one, and the performance can be further improved by combining ConPVP with the above studies.

% utilize a more efficient method to find prototypes for sentence embeddings, instead of using cluster algorithms or label information.
% However, these methods optimize PLMs by instance-wise contrastive loss, resulting in a well-separated but locally smooth embedding space.

% and those samples are constructed by rules with the syntax tree as a reference.

\begin{figure*}[!th]
\centering
\includegraphics[width=1.0\linewidth]{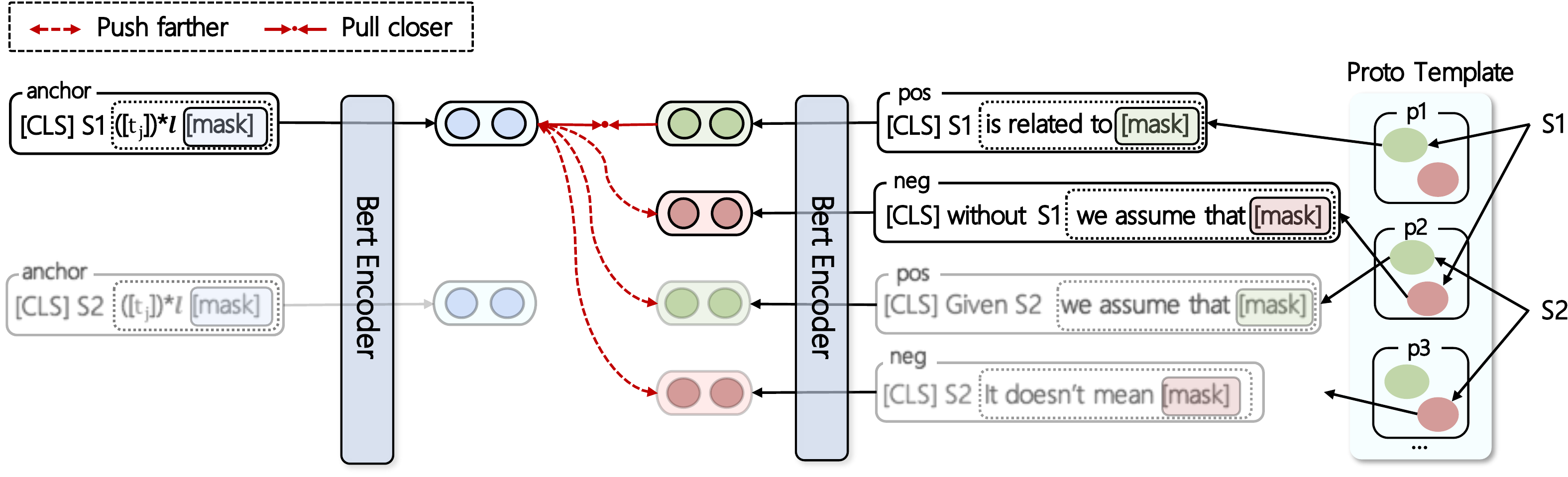}
\caption{
The overall framework of our proposed ConPVP.
}
\label{fig_model}
\end{figure*}

% construct another set of templates by manually converting each template to its negation to induce the most related negative prototypes for each sample.
% Though we construct our sentence embedding in a similar prompt-based way with the PromptBERT, when fine-tuning for sentence embedding, PromptBERT still performs contrastive learning on instance level. Unlike them, we introduce multiple prompts based on different NLP tasks to assign multi-view prototypes to each instance.
% Compared with \citet{wang2022sncse}, the XXX learn sentence embeddings in a more general way through the multi-view prototypes, which make the model capable of distinguishing the sample with high textual similarity but different semantics as well as the semantic similar one but with less words overlapping.

\section{Method}

In this section, we elaborate the proposed ConPVP, a novel contrastive learning approach implicitly encoding semantic structure into the embedding space.
As illustrated in Figure \ref{fig_model}, ConPVP is based on the popular SimCSE framework \cite{gao-etal-2021-simcse} and further leverages the concept of semantic prototypes. 
% We will describe the approach of generating {\it virtual semantic prototypes}, and then incorporate the prototypes into the contrastive learning.
% A key motivation behind this is utilizing language prompts to stimulate the knowledge of PLMs to generate semantic prototypes for each instance.

\subsection{Prompt-derived Virtual Semantic Prototypes}
% A key motivation behind this is utilizing language prompts to stimulate the knowledge of PLMs to generate semantic prototypes for each instance.

Semantic prototype is defined as a representative embedding for a group of semantically similar instances \cite{li-etal-2020-prototypical}.
% since PLMs are trained on large-scale corpora and contains a lot of rich semantic information, we use a set of manually designed templates of different downstream tasks to elicit knowledge from the PLMs to form semantic prototypes.
% On key challenge is to find and choose templates of downstream tasks.
Given the fact that PLMs are able to perform well on various NLP tasks when provided with suitable task-specific templates \cite{Brown-etal-2020-languagemodelasfewshotlearner,Sanh-etal-2021-multiprompt, Wei-etal-2021-fintuned-language-models-are-zero-shot-learners},
we can induce the semantic prototypes of sentences from PLMs with the help of prompts.
% To ensure the generalization of the prompt-derived semantic prototypes, 
In this work, we construct a template set $\mathcal{T}^+$ using four NLP tasks (i.e., classification, summarization, natural language inference, and sentence embedding), and assign each task 2 templates.
Please note that we select the templates without deliberation, and we leave the other choices as future work. 
% for each task to construct a template set $\mathcal{T}^+$.
% Based on $\mathcal{T}^+$, we further construct another template set $\mathcal{T}^-$, in which each template is modified to its negative form.
Furthermore, we construct another template set $\mathcal{T}^-$, in which the templates are the negative form of those in $\mathcal{T}^+$ and possibly induces semantically opposite response from PLMs.
% and based on $\mathcal{T}^+_D$ we further  , in which the templates are modified based the templated  manually.
% divide it into a positive template set $\mathcal{T}^+_D$ and a negative template set $\mathcal{T}^-_D$. 
All the templates are illustrated in Table \ref{tab_template}.
% different from previous studies that pre-defined a set of learnable prototype embeddings, the virtual semantic prototypes in our framework are induced by different task-specific prompts, which are not limited in number.
% Please note that  use can use more templates ...
% why this is proto

\begin{table}[!t]
\centering
\small
\begin{spacing}{1.5}
\begin{tabular}{c}
\toprule
{\bf Basic Templates}  \\
Given ``<S>'' , we assume that ``[MASK]'' \\
`` <S> '' , is this review positive ? [MASK] . \\
`` <S> '' , is [MASK] news \\
`` <S> '' , is a [MASK] one \\
`` <S> '' . In summary : `` [MASK] '' \\
By `` <S> '' they mean [MASK] . \\
Article `` <S> '' belongs to a [MASK] topic \\
This sentence : `` <S> '' means [MASK] . \\
\midrule 
{\bf Semantically Opposite Templates} \\
`` <S> '' , is this review negative ? [MASK] . \\
Without `` <S> '' , they mean [MASK] . \\
`` <S> '' is inconsistent with `` [MASK] '' \\
`` <S> '' is totally different from : `` [MASK] '' \\
`` <S> '' which does not denote [MASK] \\
`` <S> '' is not a [MASK] one \\
This sentence : `` <S> '' does not mean [MASK] . \\
Article `` <S> '' is definitely not about the [MASK] topic \\
\bottomrule 
\end{tabular}
\end{spacing}
\caption{
\label{tab_template}
{Templates for inducing semantic prototypes}.
}
\end{table}

After obtaining the template sets, we convert an input sentence $x_i$ to 
% insert template $\mathcal{T}_s$ into $X$ to convert it into the corresponding input which has a ${\rm [MASK]}$ token in it, i.e., 
$\hat{x}_i$=[$x_i$; $\mathcal{T}^{+}_i$], where $\mathcal{T}^{+}_i$ is a template sampled from $\mathcal{T}^{+}$.
We feed $\hat{x}_i$ to a PLM and take the hidden state of the ``[MASK]'' token $h_{\rm [MASK]}$ as the positive prototypical embedding $c_i^+$.
% For each sentence $x_i$, we randomly sample one template $\mathcal{T}^+_j \in \mathcal{T}^+_D$, and obtain the semantic prototypical embedding $c^+_i$.
In this same way, we generate the negative prototypical embedding $c^-_i$ using a sampled template $\mathcal{T}^{-}_i \in \mathcal{T}^{-}$.
Notably, 
unlike the conventional prototypes \cite{li-etal-2020-prototypical, Caron-etal-2020-contrasting-cluster-assignments, Sharma-etal-2020-clustering}, our method of obtaining prototypes may not seem intuitive, since there is no explicit partitioning of the embedding space. 
In order to distinguish our method from the previous studies, we name the prototypes in this work virtual prototypes.
% our prototypes are not obtained by the clustering algorithms but the task specific prompts, and we do not explicitly 
% compared to the studies obtaining prototypes by clustering algorithms \cite{li-etal-2020-prototypical, Caron-etal-2020-contrasting-cluster-assignments, Sharma-etal-2020-clustering}, 
% our proposed induce prototypes by using task specific templates, which are more flexible and generalizable, thus named virtual prototypes.

\subsection{Prototypical Contrastive Learning}

% \citet{jiang2022promptbert} propose to bring the sentence embeddings task into the prompt learning paradigm \cite{Brown-etal-2020-languagemodelasfewshotlearner}, and demonstrate its advantage in eliminating embedding biases.
% Formally, we denote $\mathcal{M}$ and $\mathcal{T}$ as a pretrained language model (e.g., BERT) and a template set.

To obtain the embedding of an anchor sentence $x_i$, 
% we first append a continuous template consisting of multiple trainable embeddings $\{t_j\}$ to $x_i$ 
% $[x_i;t_1;...;t_l;${\rm [MASK]}$]$, 
% where $l$ is the length of the continuous template, and then 
we feed $[x_i;t_1;...;t_l;${\rm [MASK]}$]$ to a PLM to obtain its contextualized representations, where $t_1;...;t_l$ is a continuous prompt.
The representation vector of the ``[MASK]'' token is taken as the sentence embedding $v_i$.
% The idea behind prototypical contrastive learning \cite{} is pulling the instance close to its semantic prototypes instead of other instances and pushing apart the instance and its irrelevant prototypes. 
% This alleviates the issues of instance-wise contrastive learning (e.g., local smoothness), thus making the embedding space more structural.
Given the embedding of the anchor sentence and the corresponding positive and negative prototype embeddings, 
% For each sentence $X_i$, we randomly sample one template $\mathcal{T}^+_i \in \mathcal{T}^+_D$, and obtain the semantic prototypical embedding $c^+_i$ using the method in \ref{promptse}.
% Similarly, we get another prototypical embedding  $c^-_i$ with a sampled template $\mathcal{T}^i_i \in \mathcal{T}^i_D$. $c^-_i$.
% wrap $X_i$ into $\widetilde{X}^+_i$ and $\widetilde{X}^-_i$.
% We take the final representations of the {\rm [MASK]} token as the semantic prototypical embedding $c^+_i$ and the converted prototypical embedding $c^-_i$.
% For each sentence $X_i$, we randomly sample one template $\mathcal{T}^+_i \in \mathcal{T}^+_D$ and one template $\mathcal{T}^-_i \in \mathcal{T}^-_D$, and wrap $X_i$ into $\widetilde{X}^+_i$ and $\widetilde{X}^-_i$.
% We take the final representations of the {\rm [MASK]} token as the semantic prototypical embedding $c^+_i$ and the converted prototypical embedding $c^-_i$.
% and then concatenate $X$ with it and obtain the sentence embedding as positive sample $h^+$. 
% Given an input $x$, template function of  $T$ 
we integrate them into the InfoNCE based contrastive loss \footnote{In practice, in order to reduce the influence of the template, we follow \cite{jiang2022promptbert} to use debiased sentence embeddings during training. Please ref \cite{jiang2022promptbert} for more detailed.}:
\begin{equation}
    l_i=-\log \frac{e^{{\rm sim}(v_{i}, c^{+}_{i})/\tau}}{\sum^{N}_{k=1} (e^{{\rm sim}(v_{i}, c^{+}_{k})/\tau} + e^{{\rm sim}(v_{i}, c^{-}_{k})/\tau})}
\end{equation}
where $N$ is the number of sentences in a mini-batch.
With this loss function, we pull the embedding of the anchor sentence $v_{i}$ close to its positive prototypical embedding $c^{+}_{i}$, and push $v_{i}$ and the irrelevant prototypical embeddings apart.
% In this way, an embedding is forced to be close to the semantic prototype to which it belongs and away from other prototype.

\begin{table}[!t]
\centering
\small
\begin{spacing}{1.3}
\begin{tabular}{lcc}
\toprule
{\bf PLM} & {\bf Batch Size} & {\bf Learning Rate} \\
\midrule
{\bf BERT-base} & 128 & 3e-5 \\
{\bf BERT-large} & 128 & 1e-5 \\
{\bf RoBERTa-base} & 128 & 1e-5 \\
{\bf RoBERTa-large} & 256 & 1e-5 \\
\bottomrule 
\end{tabular}
\end{spacing}
\caption{
\label{tab_settings_sts}
{\bf Training Settings for STS}.
% The average spearman correlation as well as the training steps of different method with different batch size.
}
\end{table}

\begin{table*}[!t]
\centering
\small
\begin{spacing}{1.25}
\begin{tabular}{lcccccccc}
\toprule
{\bf Model} & {\bf STS12} & {\bf STS13} & {\bf STS14} & {\bf STS15} & {\bf STS16} & {\bf STS-B} & {\bf SICK-R} & {\bf Avg.} \\
\midrule
\multicolumn{9}{c}{\it BERT-base} \\
% CT {\dag} & 61.63 & 76.80 & 68.47 & 77.50 & 76.48 & 74.31 & 69.19 & 72.05 \\
ConSERT {\dag} & 64.64 & 78.49 & 69.07 & 79.72 & 75.95 & 73.97 & 67.31 & 72.74 \\
SimCSE {\dag} & 68.40 & 82.41 & 74.38 & 80.91 & 78.56 & 76.85 & 72.23 & 76.25 \\
ESimCSE {\dag} & \textbf{73.40} & 83.27 & 77.25 & 82.66 & 78.81 & 80.17 & 72.30 & 78.27 \\
PromptBERT {\dag} & 71.56 & 84.58 & 76.98 & \textbf{84.47} & \textbf{80.60} & \textbf{81.60} & 69.87 & 78.54 \\
% \multicolumn{9}{l}{\bf ConPVP} \\
% \ \ w/ manual & 73.05 & \textbf{85.14} & 77.26 & 84.06 & 79.02 & 80.41 & 73.02 & \textbf{78.85} \\ 
% \ \ w/ continuous 
{\bf ConPVP} & 71.72 & \textbf{84.95} & \textbf{77.68} & 83.64 & 79.76 & 80.82 & \textbf{73.38} & \textbf{78.85} \\
\midrule
\multicolumn{9}{c}{\it BERT-large} \\
ConSERT {\dag} & 70.69 & 82.96 & 74.13 & 82.78 & 76.66 & 77.53 & 70.37 & 76.45 \\
SimCSE {\dag} & 70.88 & 84.16 & 76.43 & 84.50 & 79.76 & 79.26 & 73.88 & 78.41 \\
% & 72.86 & 83.99 & 75.62 & 84.77 & 81.80 & 81.98 & 71.26 & 78.90 \\
ESimCSE {\dag} & \textbf{73.21} & 85.37 & 77.73 & 84.30 & 78.92 & 80.73 & \textbf{74.89} & 79.31 \\
PromptBERT & 71.55 & \textbf{86.83} & \textbf{78.63} & 85.10 & 79.79 & \textbf{82.20} & 72.19 & 79.47 \\
% \multicolumn{9}{l}{\bf ConPVP} \\
% \ \ w/ manual & 72.94 & 86.66 & 78.20 & \textbf{85.56} & 79.91 & \textbf{82.44} & 74.32 & 80.00 \\
% \ \ w/ continuous &
{\bf ConPVP} & 72.63 & 86.68 & 78.14 & \textbf{85.50} & \textbf{80.13} & 82.18 & 74.79 & \textbf{80.01} \\

% & 74.75 & 84.09 & 77.88 & 83.13 & 83.44 & 83.64 & 74.31 & 80.18 \\
\midrule
\multicolumn{9}{c}{\it RoBERTa-base} \\
SimCSE {\dag} & 70.16 & 81.77 & 73.24 & 81.36 & 80.65 & 80.22 & 68.56 & 76.57 \\
ESimCSE {\dag} & 69.90 & 82.50 & 74.68 & 83.19 & 80.30 & 80.99 & 70.54 & 77.44 \\
PromptBERT {\dag} & \textbf{73.94} & \textbf{84.74} & \textbf{77.28} & \textbf{84.99} & \textbf{81.74} & 81.88 & 69.50 & 79.15 \\
% \multicolumn{9}{l}{\bf ConPVP} \\
% \ \ w/ manual & 73.92 & 84.56 & 77.08 & 84.43 & 81.44 & \textbf{82.47} & \textbf{73.89} & \textbf{79.68} \\
% \ \ w/ continuous 
{\bf ConPVP} & 73.20 & 83.22 & 76.24 & 83.37 & 81.49 & \textbf{82.18} & \textbf{74.59} & \textbf{79.18} \\
\midrule
\multicolumn{9}{c}{\it RoBERTa-large} \\
SimCSE {\dag} & 72.86 & 83.99 & 75.62 & 84.77 & 81.80 & 81.98 & 71.26 & 78.90 \\
EsimCSE {\dag} & 73.20 & 84.93 & 76.88 & 84.86 & 81.21 & 82.79 & 72.27 & 79.45 \\
PromptBERT & 72.89 & \textbf{86.44} & \textbf{78.10} & \textbf{85.09} & 79.37 & 81.52 & 70.85 & 79.18 \\
% \multicolumn{9}{l}{\bf ConPVP} \\
% \ \ w/ manual & 72.61 & 82.31 & 77.19 & 84.85 & 81.48 & 82.31 & 73.50 & 79.18 \\
% \ \ w/ continuous 
{\bf ConPVP} & \textbf{74.75} & 84.09 & 77.88 & 83.13 & \textbf{83.44} & \textbf{83.64} & \textbf{74.31} & \textbf{80.18} \\
\bottomrule 
\end{tabular}
\end{spacing}
\caption{
\label{tab_results_sts_best}
{\bf Experimental results on unsupervised STS tasks}.
Methods with {\dag} denote that we directly report the scores from corresponding paper, and others are from our implementation.
We run 4 times with different random seeds and report the best \textbf{Avg.} for fair comparison.
}
\end{table*}

\section{Experiments}

To verify the effectiveness of our proposed method, we conduct experiments and empirical analysis on Semantic Textual Similarity (STS) tasks under the unsupervised setting.

% Then, we conduct comprehensive empirical studies to better understand the advantages of our XXX.
% In this section, in order to extremly explore the effectiveness of our InforCSE, we use SimCSE \cite{} as the baseline method and conduct comprehensive empirical experiments on STS tasks, Transfer tasks and Short Text Clustering tasks, respectively.
% Then, we conduct comprehensive empirical analysis to better understand our ConPVP.

\subsection{Settings}
% \paragraph{Semantic Textual Similarity Tasks.} 
Following \citet{gao-etal-2021-simcse}, we conduct experiments on 7 semantic textual similarity (STS) tasks, including STS 2012-2016 \cite{agirre-etal-2012-semeval,agirre-etal-2013-sem,agirre-etal-2014-semeval, agirre-etal-2015-semeval, agirre-etal-2016-semeval}, STS Benchmark \cite{cer-etal-2017-semeval}, and SICK-R \cite{Marelli-etal-2014-sick}.
The similarity scores of sentence pairs in these datasets are labeled from 0 to 5.
% The relevance between gold annotations and the scores predicted by sentence vectors is measured in Spearman correlation.
% The sentence pairs in each datasets are scored from 0 to 5 to indicate semantic similarity.
% Our framework is implemented through hugging face transformers.
% \subsection{Implementation Details}
Our implementation is based on SimCSE \footnote{https://github.com/princeton-nlp/SimCSE} \cite{gao-etal-2021-simcse},
% , and we use the pretrained models provided by HuggingFace \footnote{https://github.com/huggingface/transformers} including 
and we take BERT-base \cite{devlin-etal-2019-bert}, BERT-large \cite{devlin-etal-2019-bert}, RoBERTa-base \cite{liu2019roberta}, and RoBERTa-large \cite{liu2019roberta} as our backbones\footnote{https://github.com/huggingface/transformers}.
All our experiments are conducted on
% computation node with 
a NVIDIA V100 GPU.
We set the length of the continuous prompt as 4.
Following previous studies \cite{gao-etal-2021-simcse, wu-etal-2021-esimcse}, we use 1 million sentences randomly sampled from English Wikipedia as training sentences.
We train 1 epoch, and evaluate every 125 steps and choose model parameters with highest performance on STS-B development set.
The batch size and learning rate are listed in Table \ref{tab_settings_sts}.
% More training details can be found in Appendix \ref{Appendix_STS}.

\begin{table*}[!th]
\centering
\small
\begin{spacing}{1.3}
\setlength{\tabcolsep}{1.5mm}{
\begin{tabular}{lcccccccc}
\toprule
{\bf Model} & {\bf STS12} & {\bf STS13} & {\bf STS14} & {\bf STS15} & {\bf STS16} & {\bf STS-B} & {\bf SICK-R} & {\bf Avg.} \\
\midrule
\multicolumn{9}{c}{\it RoBERTa-large} \\
% \multicolumn{9}{l}
SimCSE & 69.36$_{\pm 1.16}$ & 82.39$_{\pm 0.70}$ & 74.33$_{\pm 1.06}$ & 83.03$_{\pm 1.34}$ & 81.19$_{\pm 0.45}$ & 81.10$_{\pm 0.77}$ & 70.17$_{\pm 0.91}$ & 77.37$_{\pm 0.88}$ \\
PromptBERT & 72.00$_{\pm 1.16}$ & 83.54$_{\pm 1.85}$ & 77.05$_{\pm 1.02}$ & 83.32$_{\pm 1.15}$ & 80.82$_{\pm 0.98}$ & 82.54$_{\pm 0.79}$ & 70.31$_{\pm 0.82}$ & 78.51$_{\pm 0.77}$ \\
% \multicolumn{9}{l}{\bf ConPVP} \\
{\bf ConPVP} & \textbf{74.57}$_{\pm 0.54}$ & \textbf{83.62}$_{\pm 0.59}$  & \textbf{77.77}$_{\pm 0.29}$ & 83.18$_{\pm 0.85}$ & \textbf{82.85}$_{\pm 0.37}$  & \textbf{82.85}$_{\pm 0.46}$ & \textbf{74.47}$_{\pm 0.44}$  & \textbf{79.90}$_{\pm 0.32}$ \\
\ \ w/ manual & 72.61$_{\pm 1.26}$ & 82.31$_{\pm 0.72}$ & 77.19$_{\pm 0.55}$ & \textbf{84.85}$_{\pm 1.00}$ & 81.48$_{\pm 0.44}$ & 82.31$_{\pm 0.48}$ & 73.50$_{\pm 0.54}$ & 79.18$_{\pm 0.49}$ \\
\ \ w/o $c^-$ & 73.80$_{\pm 0.93}$ & 82.38$_{\pm 0.58}$ & 76.72$_{\pm 0.50}$ & 82.53$_{\pm 0.77}$  & 81.94$_{\pm 0.54}$  & 82.32$_{\pm 0.31}$  & 69.75$_{\pm 0.58}$ & 78.49$_{\pm 0.33}$\\
\ \ w/o $c^+$ \& $c^-$ & 72.81$_{\pm 0.75}$ & 81.51$_{\pm 0.63}$  & 74.94$_{\pm 0.77}$ & 79.83$_{\pm 0.81}$ & 80.50$_{\pm 0.55}$  & 81.06$_{\pm 0.40}$  & 70.30$_{\pm 0.43}$  & 77.28$_{\pm 0.25}$ \\
\bottomrule 
\end{tabular}}
\end{spacing}
\caption{
\label{tab_results_ablation_study}
{\bf Ablation Study}.
We run each experiment 4 times with different random seeds and report mean and standard deviation.
}
\end{table*}

\subsection{Main Results}

We compare our ConPVP to the recent related methods which are based on instance-wise contrastive learning, including 1) {\it ConSERT} \cite{yan-etal-2021-consert} which exploits four data augmentation strategies to construct positive samples;
2) {\it SimCSE} \cite{gao-etal-2021-simcse} which directly uses Dropout to generate positive pairs;
3) {\it ESimCSE} \cite{wu-etal-2021-esimcse} which introduces word repetition augmented positive pairs and momentum negative pairs;
4) {\it PromptBERT} \cite{jiang2022promptbert} which reformulates the sentence embeddings task as a prompt-based learning paradigm.
% We denote our ConPVP with manual-designed prompt as ConPVP w/ manual and our method with continuous prompt as ConPVP w/ continuous, respectively.

\begin{figure*}[!th]
\centering
\includegraphics[width=1.0\linewidth]{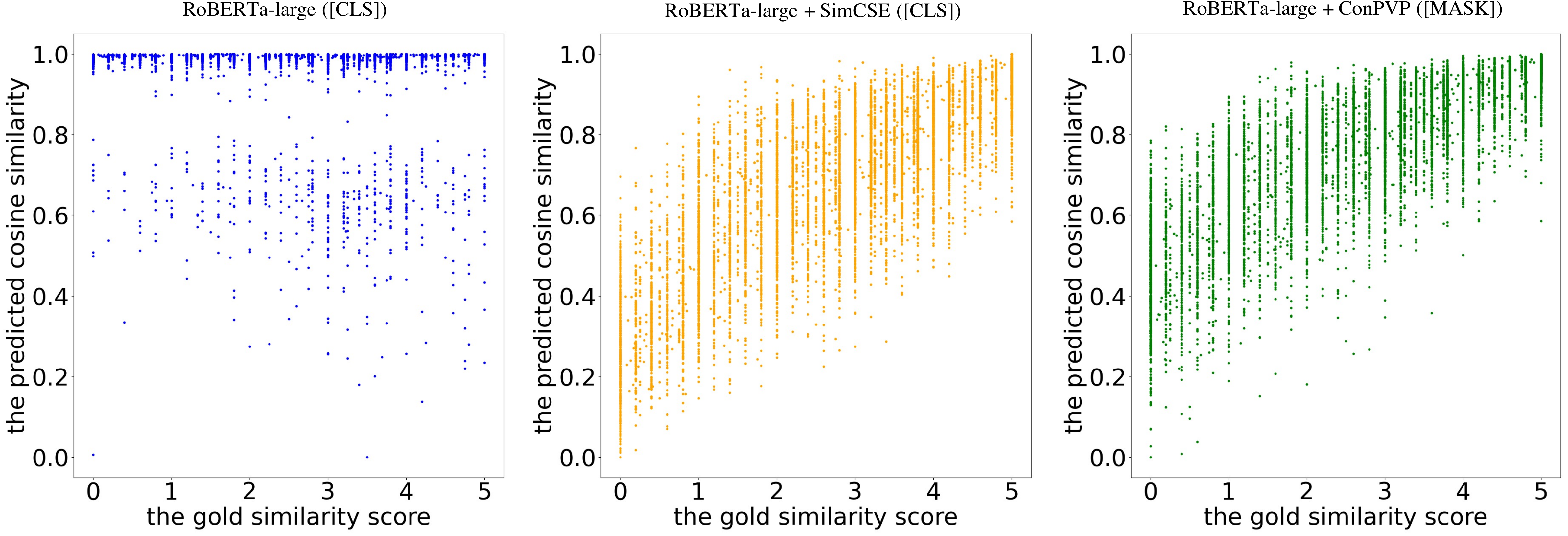}
\caption{
{\bf Distribution of predicted cosine similarity.}
The correlation diagram between the gold similarity scores (x-axis) and model predicted cosine similarity scores (y-axis) on the STS-B dataset.
% with the RoBERTa-large as the backbone.
We scale the predicted scores to 0 to 1.
}
\label{fig_distribution}
\end{figure*}

For fair comparison, we report the best performance from 4 runs in Table \ref{tab_results_sts_best}.
Compared with {\it SimCSE}, {\it ConPVP} brings significant improvements across the board.
% Specifically, ConPVPs w/ manual achieve 2.60 points, 1.59 points, 3.11 points and 0.28 points (on Avg.) improvement over BERT-base, BERT-large, RoBERTa-base and RoBERTa-large, respectively.
Specifically, 
{\it ConPVP} achieves average improvements of 2.60, 1.60, 2.61, and 1.28 points over BERT-base, BERT-large, RoBERTa-base and RoBERTa-large, respectively, showing the superiority of our prototypical contrastive method.
Besides, 
{\it ConPVP} surpasses {\it ConSERT} and {\it ESimCSE}, which carefully design positive samples with various textual data augmentation.
This demonstrates that although the textual data augmentation can provide different views of the anchor, these methods based on it still suffers the local smooth problem. 
In contrast, our model shows that textual data augmentation is possibly unnecessary, and the improvement can be achieved by encoding more structural information into the embedding space, e.g., finding semantic prototypes.
Finally, our {\it ConPVP} achieves consistently better performance than {\it PromptBERT}, demonstrating the effectiveness of the proposed prototypical contrastive loss.

% Finally, it is worth noting that {\it ConPVP w/ continuous} yield competitive or even better results against {\it ConPVP w/ manual}, where the manual prompt is carefully searched by \cite{jiang2022promptbert}.
% This may because one manually-designed prompt cannot be the optimal choice for all PLMs at the same time, while the continuous prompt is more flexible to different PLMs.
% Thus, we recommend use continuous prompt in our framework, and we conduct analysis on ConPVP w/ continuouse in the following sections to explore the advantages of our proposed method.
% In particular, PromptBERT performs worse than SimCSE in SICK-R, while our ConPVP outperforms {\it SimCSE} by a large margin on SICK-R.
% The reason may be that SICK-R is essentially a different task from others, which is constructed by natural language inference sentences.
% It focuses specifically on the ability of PLMs to model semantic similarity. 
% These observations verify our claim that the contrastive learning methods with textual data augmentation positive samples fail to decouple the textual similarity and semantic similarity.
% And, the prompt-derived informative samples in our framework benefit the modeling of semantic similarity.

\subsection{Ablation Study}

To analyze the impact of different components of ConPVP, we investigate the following three variants: 1) {\it ConPVP w/ manual}, where we obtain the anchor sentence embeddings with the searched discrete templates from \citet{jiang2022promptbert};
2) {\it ConPVP w/o $c^-$}, where we remove the negative prototypes in the prototypical contrastive loss; 
3) {\it ConPVP w/o $c^+$ \& $c^-$}, which is equivalent to SimCSE but uses continuous prompts for sentence embeddings.
% in which we remove all of the prototypes and only keep the continuous prompts used to induce the anchor embeddings.
Notably, {\it ConPVP w/o $c^+$ \& $c^-$} is also a variant of PromptBERT \cite{jiang2022promptbert}, where the discrete templates are replaced by the continuous ones.
% we formulate sentence embedding task as continuous prompt based learning and optimize the model with traditional instance-based contrastive loss.
We take RoBERTa-large as the backbone.

\begin{figure}[!th]
\centering
\includegraphics[width=1.0\linewidth]{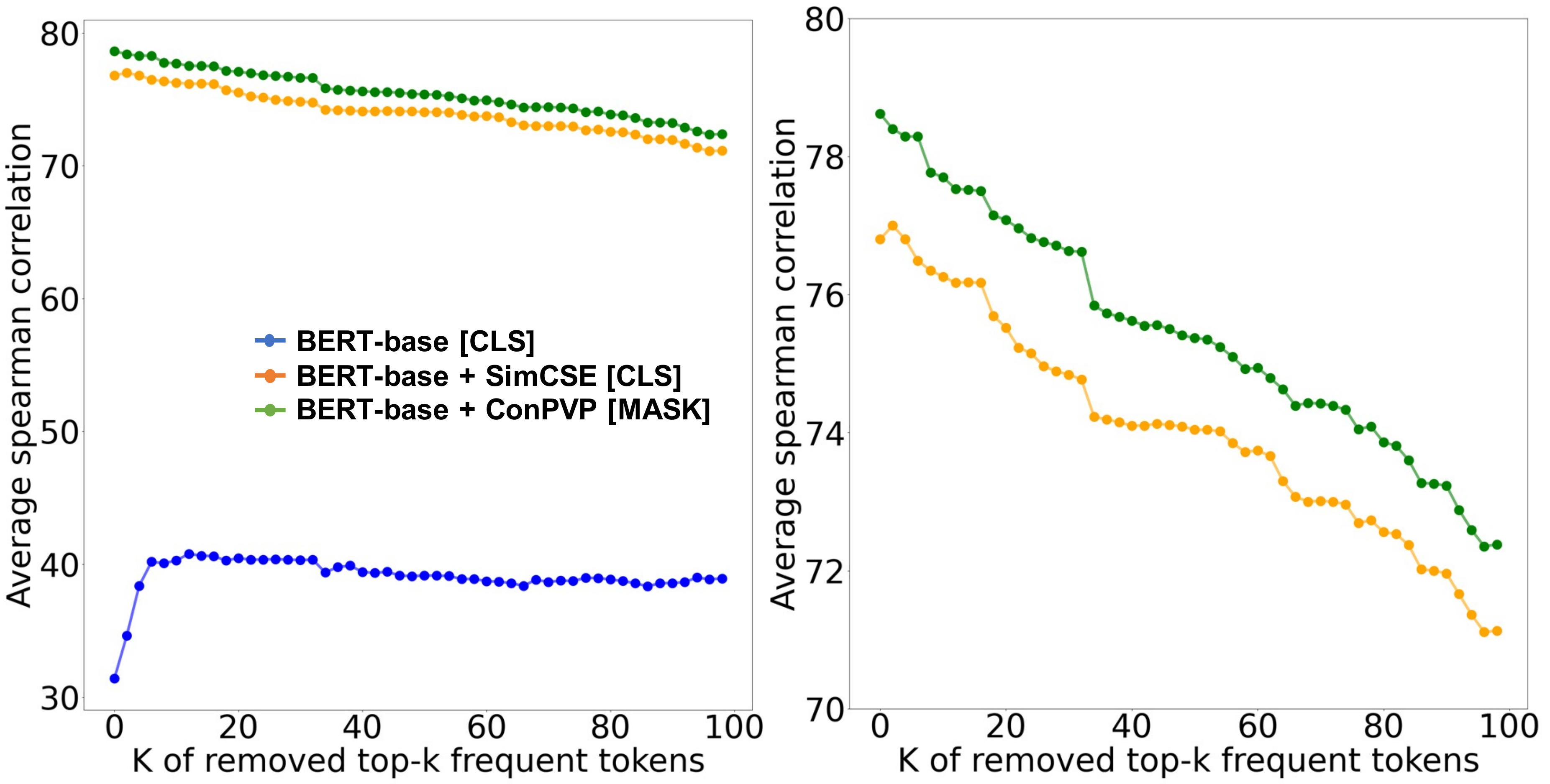}
\caption{
{\bf Analysis of embedding space.} The average spearman correlation on STS tasks w.r.t the number of removed top-k frequent tokens.
The frequency of each token is calculated through the test split of the STS Benchmark dataset.
}
\label{fig_remove_k}
\end{figure}

The results on STS tasks are listed in Table \ref{tab_results_ablation_study} and the conclusions are as follows:
1) {\it ConPVP} obtains better results against {\it ConPVP w/ manual}.
% , where the manual prompt is carefully searched by \cite{jiang2022promptbert}.
This may be because one manually-designed prompt cannot fit different PLMs and training strategies at the same time, and continuous prompts are more flexible and effective in comparison.
Besides, the improvement of {\it ConPVP w/ manual} over {\it PromptBERT} validates the advantage of the prototypical contrastive loss.
2) Removing the negative prototypes (i.e., {\it ConPVP w/o $c^-$}) leads to a performance degradation of 1.41 point against {\it ConPVP}.
The underlying reason is that the negative prototypes here serve as a type of hard negatives—the semantics of the negative prototypes are essentially different from the positive prototypes but the prompts used to induce them are similar in text.
% essentially different to the anchor sentence in semantics,
% despite textually similar to the anchor sentence but have essentially different semantics.
% The performance drop of {\it ConPVP w/o $c^-$} compared with {\it ConPVP}
% and ConPVP w/o $c^+$ \& $c^-$ 
% indicates the effectiveness of the introduced negative prototypes.
% In particular, removing the negative prototypes leads to a significant performance degradation of 4.72 point on SICK-R, which focuses on semantic similarity (e.g., entailment, contrastive, or natural) instead of textual similarity. 
% Our negative prototypes have essentially different semantics from the anchor sentence.
% Incorporate them into contrastive learning benefits the modeling of semantic similarity. 
3) We find that {\it ConPVP w/o $c^+$ \& $c^-$} does not give an improvement against {\it SimCSE}.
% and {\it ConPVP} significantly outperforms {\it SimCSE} and {\it ConPVP w/o $c^+$ \& $c^-$}.
These observations show that the gain of our method entirely comes from the cooperation between the prompt-derived virtual prototypes and the prototypical contrastive loss, rather than the usage of the prompt-based sentence embeddings.

\subsection{Distribution of Cosine Similarity}
In this section, we investigate the similarity distributions learned by different methods.
As shown in Figure \ref{fig_distribution}, 
the native sentence representations of {\it RoBERTa-large} suffer from the collapse issue \cite{Chen-etal-2021-exploring}, and therefore we get high similarity scores for all sentence pairs.
By contrast, both {\it ConPVP} and {\it SimCSE} alleviate the collapse issue, and the predicted cosine similarity scores for positive pairs of {\it ConPVP} are more certain.
For example, for the positive pairs whose similarity scores range from 4 to 5, the scores predicted by {\it ConPVP} (0.6 to 1.0) is more concentrated than the scores predicted by {\it SimCSE} (0.4 to 1.0).

\begin{figure*}[!th]
\centering
\includegraphics[width=1.0\linewidth]{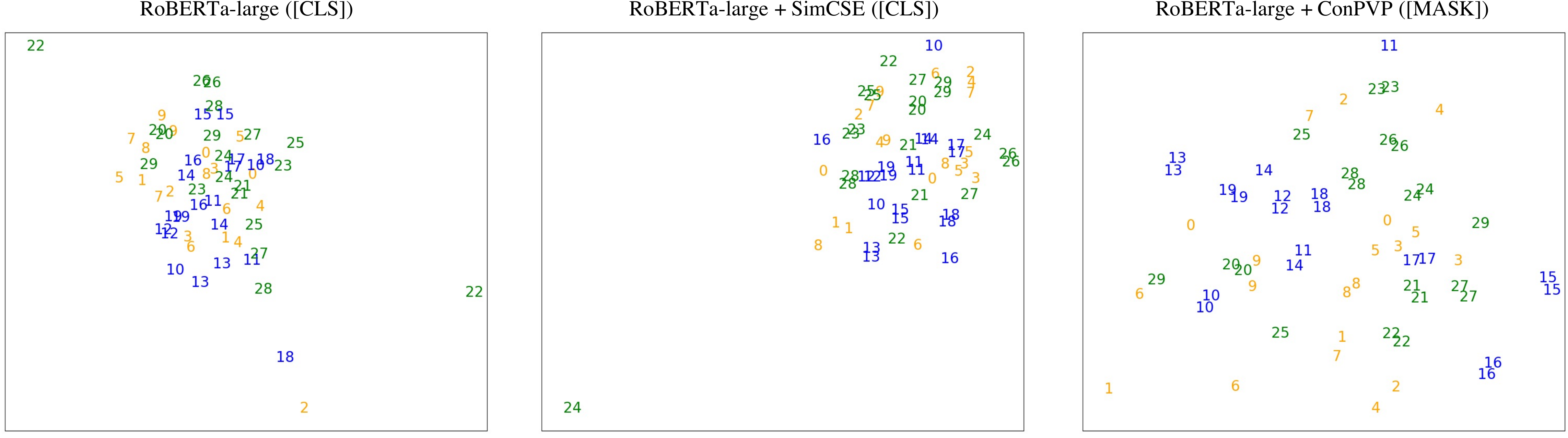}
\caption{
{\bf Visualization of learned embeddings.} 
We visualize 10 sentence pairs whose similarity scores are 0 in {\color{orange} orange} (ids from 0 to 9), 10 pairs whose similarity scores are 3 in {\color{blue} blue} (ids from 10 to 19), and 10 pairs whose similarity scores are 5 in {\color{green} green} (ids from 20 to 29).
The sentences are sampled from the STS-B test set.
}
\label{fig_visualization}
\end{figure*}

\begin{table*}[!th]
\centering
\small
\begin{spacing}{1.3}
\setlength{\tabcolsep}{1.5mm}{
\begin{tabular}{lcccccccc}
\toprule
{\bf Model} & {\bf MR} & {\bf CR} & {\bf SUBJ} & {\bf MPQA} & {\bf SST} & {\bf TREC} & {\bf MRPC} & {\bf Avg.} \\
\midrule
GloVe {\dag} & 77.25 & 78.30 & 91.17 & 87.85 & 80.18 & 83.00 & 72.87 & 81.52 \\
Skip-thought {$^\heartsuit$} & 76.50 & 80.10 & 93.60 & 87.10 & 82.00 & 92.20 & 73.00 & 83.50 \\
IS-BERT {$^\heartsuit$} & 81.09 & 87.18 & 94.96 & 88.75 & 85.96 & 88.64 & 74.24 & 85.83 \\
\midrule
\multicolumn{9}{c}{\it RoBERTa-base} \\
SimCSE {\dag} & 81.04 & 87.74 & 93.28 & 86.94 & 86.60 & 84.60 & 73.68 & 84.84 \\
SimCSE & 81.75$_{\pm 0.19}$ & 87.23$_{\pm 0.08}$ & 93.18$_{\pm 0.13}$ & 87.13$_{\pm 0.06}$ & 86.98$_{\pm 0.39}$ & 85.40$_{\pm 0.71}$ & 73.78$_{\pm 0.11}$ & 85.06$_{\pm 0.13}$ \\
Ours & \bf 82.44$_{\pm 0.17}$ & \bf 88.30$_{\pm 0.16}$ & \bf 93.20$_{\pm 0.11}$ & \bf 88.74$_{\pm 0.06}$ & \bf 87.70$_{\pm 0.07}$ & \bf 87.33$_{\pm 0.25}$ & \bf 76.15$_{\pm 0.19}$ & \bf 86.27$_{\pm 0.11}$ \\
\midrule 
\multicolumn{9}{c}{\it RoBERTa-large} \\
SimCSE {\dag} & 82.74 & 87.87 & 93.66 & 88.22 & 88.58 & 92.00 & 69.68 & 86.11 \\
SimCSE & 83.17$_{\pm 0.41}$ & 88.46$_{\pm 0.43}$ & 93.73$_{\pm 0.10}$ & 88.33$_{\pm 0.10}$ & 88.52$_{\pm 0.29}$ & 91.40$_{\pm 0.71}$ & 71.34$_{\pm 1.17}$ & 86.42$_{\pm 0.23}$ \\
ConPVP & \bf 85.65$_{\pm 0.28}$ & \bf 90.73$_{\pm 0.32}$ & \bf 94.13$_{\pm 0.13}$ & \bf 90.03$_{\pm 0.23}$ & \bf 89.81$_{\pm 0.29}$ & \bf 93.40$_{\pm 0.16}$ & \bf 76.47$_{\pm 0.29}$ & \bf 88.60$_{\pm 0.14}$ \\
\bottomrule 
\end{tabular}}
\end{spacing}
\caption{
\label{tab_results_transfer}
{\bf Experimental results on Transfer tasks with {\it RoBERTa-base} and {\it RoBERTa-large} backbones}.
{\dag}: results from \citet{gao-etal-2021-simcse}.
$^\heartsuit$: results from \citet{zhang-etal-2020-unsupervised}.
We run 4 times with different random seeds and report the average accuracy and standard deviation. 
}
\end{table*}

\subsection{Analysis of Embedding Space}

Previous studies indicated that the collapse issue is mainly due to anisotropy of the learned embedding space, which is sensitive to token frequency \cite{yan-etal-2021-consert, jiang2022promptbert}.
We follow \citet{yan-etal-2021-consert} to remove the embeddings of K most frequent tokens and explore the relation between the number of removed tokens and the average spearman correlation on STS tasks.

From Figure \ref{fig_remove_k}, we can observe that the performance of native {\it BERT-base} and {\it SimCSE} improves when removing the most frequent tokens.
By contrast, {\it ConPVP} achieves its best performance without removing any tokens, showing that our approach reshapes the BERT's original embedding space, reducing the influence of common tokens on sentence representations.
In addition, the performance of both {\it SimCSE} and {\it ConPVP} drops as the number of removed tokens increases but {\it ConPVP} performs significantly better, demonstrating the robustness of {\it ConPVP} to incomplete input.

\subsection{Visualization of Learned Embeddings}

We visualize a few variants of {\it RoBERTa-large} sentence embeddings to grasp an intuition on the effectiveness of our method.
Specifically, we sample 3 groups of samples from the STS-B test set, and the similarity score of each group is 0 (orange), 3 (blue), and 5 (green), respectively.
Each group has 10 sentence pairs.
We visualize their embeddings generated by different models using t-SNE \cite{maaten2008sne} in Figure \ref{fig_visualization}.

Due to the collapse issue, the sentence embeddings obtained from {\it RoBERTa-large [CLS]} cluster together whether they are similar or not.
For SimCSE, the sentence embeddings of the positive pairs are well-clustered.
However, the sentences pairs with similarity scores of 3 or 5 are very close in the embedding space.
In contrast, the embeddings learned by our ConPVP are more discriminative, forming more separated clusters (e.g., the sentence pairs in green are more clustered than those in blue, while the pairs in orange are more dispersed).
% encourages the sentence embeddings not entirely focus on pulling positive pairs together and pushing negative pairs farther, but to 
% consider the relative semantic distance between sentences, which can better benefit semantic textual similarity task.
% The phenomenon is because ConPVP measure the similarity of sentences through various semantic prototypes instead of instance discrimination.
% distance-discriminative alignment of positive sentence pairs and the negative sentence pairs are relative far apart.
% In figure \ref{}, we visualize the unsupervised learned sentence embedding of BERT-large trained with different contrastive methods using t-SNE \cite{}.
% Compared to the sentence embedding learned by SimCSE, the representation learned by the proposed ConPVP forms more separated clusters, which also suggests representation of lower entropy.

\begin{figure*}[!th]
\centering
\includegraphics[width=1.0\linewidth]{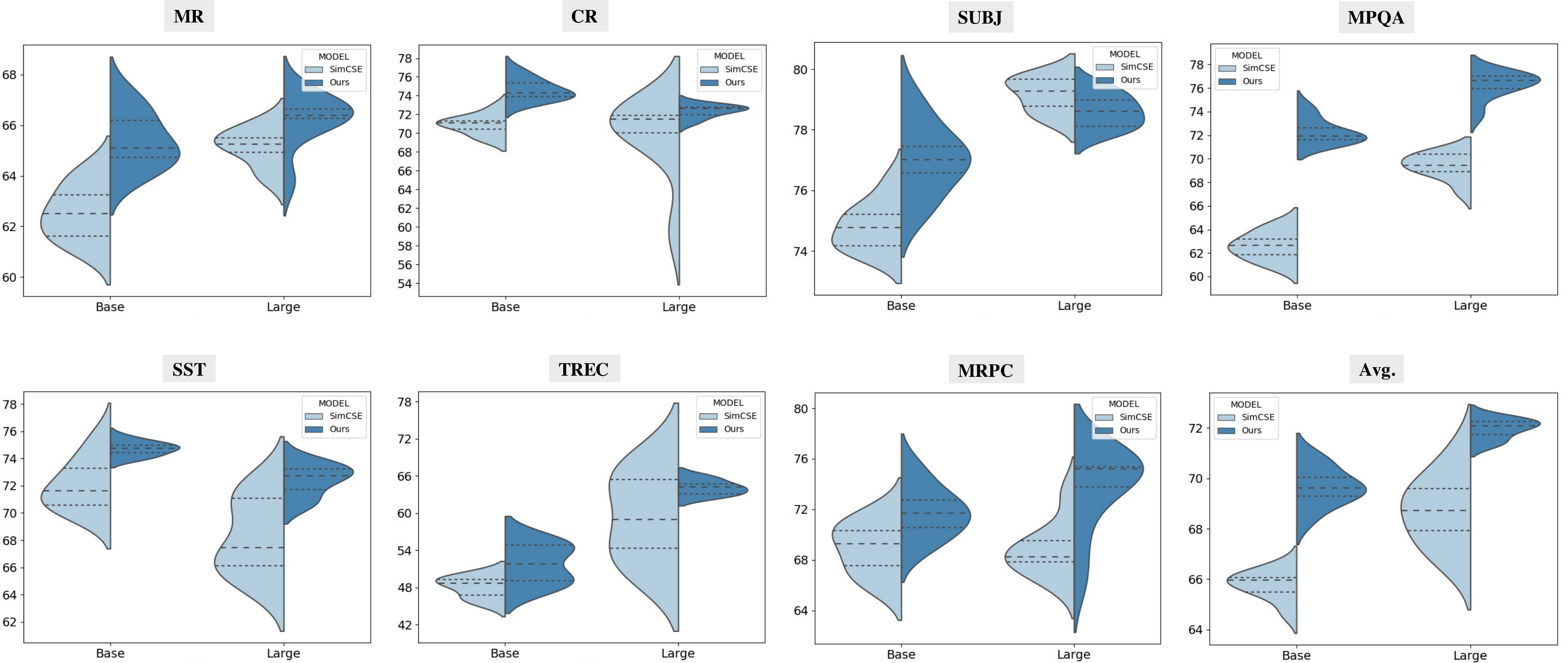}
\caption{
{\bf Few-shot learning evaluation on Transfer tasks with {\it RoBERTa-base} and {\it RoBERTa-large} as the backbones}.
For each task, we randomly sample 16 labeled instances per class and draw violin plots of the performance of 10 runs with different random seeds.
}
\label{fig_few_shot}
\end{figure*}

\begin{table*}[!th]
\centering
\small
\begin{spacing}{1.2}
\begin{tabular}{lccccccccc}
\toprule
{\bf Model} & {\bf AG} & {\bf Bio} & {\bf Go-S} & {\bf G-T} & {\bf G-TS} & {\bf SS} & {\bf SO} & {\bf Tweet} & {\bf Avg.} \\
\midrule
\multicolumn{10}{c}{\it BERT-base} \\
BERT & 79.56 & 32.46 & 54.35 & 47.12 & 61.61 & 64.04 & 21.87 & 45.35 & 50.80 \\
SimCSE {$^\clubsuit$} & 74.36 & 35.89 & 58.90 & 57.28 & 65.03 & 64.32 & 50.57 & 54.28 & 57.58 \\
\textbf{ConPVP} & \textbf{77.21} & \textbf{41.84} & \textbf{61.98} & \textbf{59.87} & \textbf{67.56} & \textbf{73.28} & \textbf{73.06} & \textbf{56.06} & \textbf{63.86} \\
\multicolumn{10}{c}{\it BERT-large} \\
BERT & \textbf{83.13} & 30.52 & 56.34 & 46.11 & 61.51 & 66.54 & 26.10 & 44.20 & 51.81  \\
SimCSE {$^\clubsuit$} & 80.23 & \textbf{43.47} & 61.87 & \textbf{61.05} & 65.78 & 68.97 & \textbf{68.03} & 55.08 & 63.06 \\
\textbf{ConPVP} & 82.50 & 41.26 & \textbf{63.82} & 58.87 & \textbf{68.34} & \textbf{74.39} & 66.59 & \textbf{57.34} & \textbf{64.14} \\
\bottomrule 
\end{tabular}
\end{spacing}
\caption{
\label{tab_results_cluster}
{\bf Clustering accuracy reported on short text clustering datasets with {\it BERT-base} and {\it BERT-large} as the backbones}.
{$\clubsuit$}: results evaluated on the checkpoints provided by \cite{gao-etal-2021-simcse}.
We report the clustering accuracy averaged over 10 independent runs.
% We run 4 times with different random seeds and report the average accuracy and the standard deviation. 
}
\end{table*}

\section{Application to Transfer Learning Tasks}

We evaluate the quality of the sentence embeddings learned by ConPVP on transfer learning tasks, including 
% We evaluate our models on the following transfer tasks:
MR \cite{Pang-etal-2005-mr}, CR \cite{Hu-etal-2004-cr}, SUBJ \cite{Pang-etal-2004-subj}, MPQA \cite{Wiebe-etal-2005-mpqa}, SST-2 \cite{Socher-etal-2013-sst}, TREC \cite{Li-etal-2002-trec} and MRPC \cite{Dolan-etal-2004-mrpc}.
A logistic regression classifier is trained using frozen sentence embeddings produced by different methods.
% We use RoBERTa-base and RoBERTa-large as the backbones.
We follow default configurations from SentEval \cite{Conneau-etal-2018-senteval}.
In addition, based on the principle that good representations can be transferred well with limited supervision and fine-tuning, we extend the evaluation to few-shot setting and follow \cite{Zhang-etal-2021-pairsupcon} to uniformly sample 16 labeled instances per class for each task.
% Detailed of data statistic and settings is provided in Appendix \ref{}.
% We run 4 times with different random seeds and report the average accuracy and the standard deviation.

% \subsection{Main Result}

% \subsection{Performance under Few-shot Settings}
% To validate the reliability and the robustness of ConPVP under the data scarcity scenarios, we conduct the few-shot experiments.
% We limit the number of unlabeled texts to 1, 10, 100, 1000, and 10000 respectively, and compare their performance with the full dataset.

Table \ref{tab_results_transfer} presents the results under the full data setting.
As we can see, the performance gap between {\it ConPVP} and {\it SimCSE} is significant and consistent.
Furthermore, we can observe more obvious gap under the few-shot setting (Figure \ref{fig_few_shot}).
The results reveal the robustness and effectiveness of our approach under the data scarcity scenarios, which is important in real-world applications.
% With only a small amount of unlabeled texts draw from the target data distribution, our approach can also tune the representation space and benefit the downstream tasks.

\section{Application to Clustering Tasks}

% Existing work mainly focuses on the semantic similarity tasks.
% Desirably, our method maps the instances from the same category close together in the representation space while maps those from different categories far apart.
% Therefore, we follow 
We follow
\citet{Zhang-etal-2021-pairsupcon} to consider 8 benchmark datasets for short text clustering, including SearchSnippets (SS) \cite{Phan-etal-2008-SS}, StackOverflow (SO) \cite{Xu-etal-2017-SO}, Biomedical (Bio) \cite{Xu-etal-2017-SO}, AgNews (AG) \cite{Zhang-etal-2015-AG}, Tweet \cite{Yin-etal-2016-Tweet} and GoogleNews (G-T, G-S, G-TS) \cite{Yin-etal-2016-Tweet}.
We follow default settings of \cite{Zhang-etal-2021-pairsupcon} and use {BERT-base} and {BERT-large} as the backbones.
We run K-Means \cite{Pedregosa-etal-2011-scikitlearn} on the sentence embeddings and report the clustering accuracy averaged over 10 independent runs.
% Detailed of datasets and settings is introduced in Appendix \ref{}.
% , including SearchSnippets (SS) \cite{}, StackOverflow (SO) \cite{}, Biomedical (Bio) \cite{}, AgNews (AG) \cite{}, Tweet \cite{} and GoogleNews (G-T, G-S, G-TS) \cite{} and use BERT-base, BERT-large as the backbones.
% We follow default settings of \cite{}.
% We run K-Means with the scikit-learn package \cite{} on the representations generated by each model and report the clustering accuracy averaged over 10 independent runs.
% \subsection{Main Result}
As illustrated in Table \ref{tab_results_cluster}, 
in comparison with {\it SimCSE}, {\it ConPVP} obtains an averaged improvement of 1.21 and 2.18, respectively, which validates our motivation in leveraging the implicit grouping effect of the prompt-derived semantic prototypes to encode more semantic structure into representations.

\section{Conclusion}
In this work, we take the first step to explore the prototypical contrastive learning on unsupervised sentence embedding learning, and consider more semantic views for each instance than the recent instance-wise contrastive methods.
In particular, we make use of the prompting in PLMs to generate the positive and negative prototypical embeddings with task-specific templates.
The experiments and extensive analysis 
% results on unsupervised semantic textual similarity tasks demonstrate the effectiveness of our method, and extensive analysis on transfer and clustering tasks also 
validate the effectiveness and robustness of our ConPVP.

\section{Limitations}
We only tried 16 task specific prompts in this paper, which is possibly sub-optimal to induce semantic prototypes.
Besides, the usage of prompts reduces the maximum effective lengths that the pretrained language models can process.

\section*{Acknowledgements}
We would like to thank anonymous reviewers for their suggestions and comments.

% % Entries for the entire Anthology, followed by custom entries
\bibliography{anthology,custom}
\bibliographystyle{acl_natbib}

\end{document}